\documentclass[sigconf,authorversion,nonacm]{acmart}
\hypersetup{pdfborder={0 0 0}}

\begin{document}

\title{Towards Illusions Awareness in Cyber-Physical System's Design}

\author{Anna Di Placido}
\email{anna.di-placido@etu.univ-cotedazur.fr}
 \affiliation{%
   \institution{Université Côte d’Azur,\\ I3S/INRIA Kairos}
   \city{Sophia Antipolis}
   \country{France}
 }

 \author{Nicolas Ferry}
 \email{nicolas.ferry@univ-cotedazur.fr}
 \affiliation{%
   \institution{Université Côte d’Azur,\\ I3S/INRIA Kairos}
   \city{Sophia Antipolis}
   \country{France}
 }

 \author{Julien Deantoni}
 \email{julien.deantoni@univ-cotedazur.fr}
 \affiliation{%
   \institution{Université Côte d’Azur,\\ I3S/INRIA Kairos}
   \city{Sophia Antipolis}
   \country{France}}

\begin{abstract}
Cyber-Physical Systems (CPS) operate through a continuous sense-compute-act loop within an open context environment, making it impossible to anticipate all the situations the system will face. To cope with this openness, stakeholders rely on assumptions, formalized into design models. However, these assumptions may no longer hold once the system is confronted with runtime reality, resulting in a discrepancy between expected and observed behaviour known in literature as the \emph{reality gap}. 
Existing approaches mainly focus on reducing or overcoming it by making simulations more faithful to reality, with no unified methodology to structure and exploit invalidated assumptions that give rise to this gap as reusable design knowledge.
We refer to the persistent reliance on invalidated assumptions -- and the resulting false confidence in the design model's operational validity -- as \emph{design illusions}, and argue that they need to be made explicit, structured, and exploited as knowledge to support better design decisions. 
We propose a conceptual pipeline for illusions-awareness that identifies, classifies, characterizes, and leverages illusions 
to transform them into actionable design knowledge.
\end{abstract}


\maketitle

\section{Introduction}
 
Cyber-Physical Systems (CPS) are increasingly deployed in domains such as aerial drones, autonomous vehicles, and smart manufacturing, where software interacts with the physical world through a continuous perceive-compute-act loop. 
This loop operates within a continuously evolving, partially observable environment, often referred to as an open context \cite{Stellet2019}.
System design relies on models and simulations to anticipate system behaviour, but these typically remain valid only under controlled conditions \cite{Acker2024}. The openness of the operational context prevents anticipating all environmental conditions and interactions the system will face. To reason about such systems, stakeholders involved in the system design and development typically rely on several mechanisms. One of them is \emph{abstraction}, which captures only the relevant aspects of the system's behaviour and its expected interactions with the environment. Another is \emph{approximation}, which provides numerical or qualitative estimates that deviate from the true system behaviour within an accepted margin of error. These mechanisms are built upon \emph{assumptions} made by stakeholders -- \textit{i.e.,} statements assumed to be true about the system and its operating context during design and analysis. To evaluate system behaviour under these assumptions, stakeholders rely on simulation-based approaches, as well as more recent techniques such as digital twins. 
These approaches enable the analysis of system performance under controlled, reproducible conditions--via simulation scenarios or, for digital twins, operational data.
Once the system reaches an acceptable level of performance with respect to its objectives under these assumptions, it may be deployed, meaning that it is released for operation in a real-world environment.

During operation, the initial assumptions made during design and validated through simulations may no longer hold. Runtime data can reveal previously unknown patterns and behaviours that were not anticipated during development, challenging or invalidating these assumptions and potentially leading to discrepancies between the expected and observed behaviour of the system.

Such discrepancy between design-time models and the behaviour observed during operation is commonly referred to as the \emph{reality gap}. This gap can emerge from multiple heterogeneous sources, \textit{e.g.,} hardware limitations \cite{Marti2019}, environmental unpredictability \cite{Broman2023}, temporal delays \cite{Lee2006}, and modelling approximations \cite{derler2012}, which may appear or accumulate at different stages of the system's life cycle. The reality gap has been widely studied in the literature, particularly in fields such as robotics, sim-to-real transfer, and digital twins, but existing approaches mainly focus on reducing or overcoming it by making simulations more faithful to reality~\cite{aljalbout2025realitygaproboticschallenges}.

We refer to the persistent reliance on invalidated assumptions--and the resulting false confidence in the design model's operational validity--as \emph{design illusions}. Existing approaches provide mechanisms to detect or mitigate discrepancies between models and reality, but there is still no unified methodology for systematically exploiting design illusions as a source of knowledge for design. 

The goal of this work is to make design illusions explicit and actionable for stakeholders by identifying and analysing them to explain observed discrepancies and to derive new knowledge that can be incorporated into design models.

The contributions of this paper are: (i) the conceptualization of invalidated assumptions as \emph{design illusions}; and (ii) an illusion-awareness conceptual pipeline to identify, classify, characterize, and leverage these illusions to improve stakeholders' awareness of the assumptions and limitations of the underlying design models.
This paper is organized as follows. Section 2 presents our motivating example. Section 3 defines how illusions may appear and accumulate during the system's life cycle and presents the illusions-awareness pipeline. Section 4 positions our proposition within related works and Section 5 concludes.

\section{Motivation example}
As a motivating example, we consider the development of an autonomous Crazyflie \cite{crazyflieGithub}, an open source nano drone, which embeds a wall-following controller, obstacle avoidance capabilities, and sensors providing distance as well as localisation information.
Its mission is to fly through an apartment by following the walls while avoiding obstacles without collisions.
During development, stakeholders rely on their knowledge of the Crazyflie's architecture, including its sensing capabilities, localisation mechanisms, and control logic, as well as on their understanding of the operating environment. They formulate the following assumptions, to successfully carry out the mission: 
\begin{itemize}
    \item \textbf{A1}: The forward-facing range sensor detects the wall before the minimum safety distance $d_{\mathrm{safe}}$ is violated;
    \item \textbf{A2}: The estimated position of the Crazyflie relative to the wall remains sufficiently accurate for the controller to preserve the safety distance.
    \item \textbf{A3}: Every obstacles detected along the planned trajectory triggers an avoidance maneuver;
\end{itemize}

The assumptions A1, A2, and A3 are then embedded into software and simulation models used during development, and specifying the\textit{expected behaviour} of the system: the Crazyflie completes its mission while maintaining the required safety distance from the wall. However, during operation, the Crazyflie repeatedly collide with the wall instead of maintaining the expected safety distance. Although no implementation faults is observed, the \textit{observed behaviour} contradicts the expected behaviour. 

Several explanations may account for this discrepancy: A1 is violated if the wall is detected too late to preserve the required safety distance; A2 is violated if localisation errors cause the controller to overestimate the distance to the wall, leading the drone to believe the safety distance is respected when it is not (\textit{i.e.}, A1 not being violated); A3 is violated if obstacles detection does not trigger the intended avoidance maneuver. 

In our case, we observed A2 is violated. The controller relies on the estimated distance provided by the localisation system rather than the actual distance to the wall. For instance, in our case, the controller was designed to maintain a safety distance $d_{\mathrm{safe}}$ of 200 mm. During operation, the localisation system reported a distance of 210 mm, while the actual distance was only 190 mm. The controller therefore considered that the safety requirement was satisfied and continued its trajectory, although the Crazyflie was already too close to the wall.

Execution traces confirm this: collisions occur whenever the gap between the estimated and the actual distances becomes significant. In the incident, the controller maintained the required 200 mm distance according to its estimation, while the real distance continuously decreased until a collision occurred. This shows that the assumption on localisation accuracy was not valid under operational conditions.
This evidence allows developers to refine A2 by explicitly modelling the expected localisation error and its impact on the safety margin. For instance, instead of assuming that the estimated distance directly represents the real distance, the controller model can be updated to consider an uncertainty interval and maintain a larger estimated distance when the localisation error increases.

Design illusions may also arise from interacting couplings: \textit{e.g.,} traces from longer missions may reveal higher-than-anticipated battery consumption due to avoidance maneuvers, requiring the original energy model to be revised.

These examples show that assumptions made during development may seem valid while becoming invalid or incomplete when the system operates in its real environment. They also illustrates (i) the role of execution traces in revealing design illusions and (ii) how by incorporating the information about the illusions obtained from execution, developers can update their models, revise assumptions, and improve the confidence in the system's behaviour.

The following section generalises the observations from these examples by defining the problem of managing design illusions and presenting a conceptual pipeline for design illusions awareness.

\section{Managing design illusions}

This section first formalizes the notion of design illusion and introduces the behavioural representations used throughout this work. It then presents a conceptual pipeline for identifying, classifying, characterizing, and leveraging design illusions in order to improve designers' awareness to make future better decisions.

\subsection{Problem definition}

\begin{figure*}
    \centering
    \includegraphics[width=0.8\linewidth]{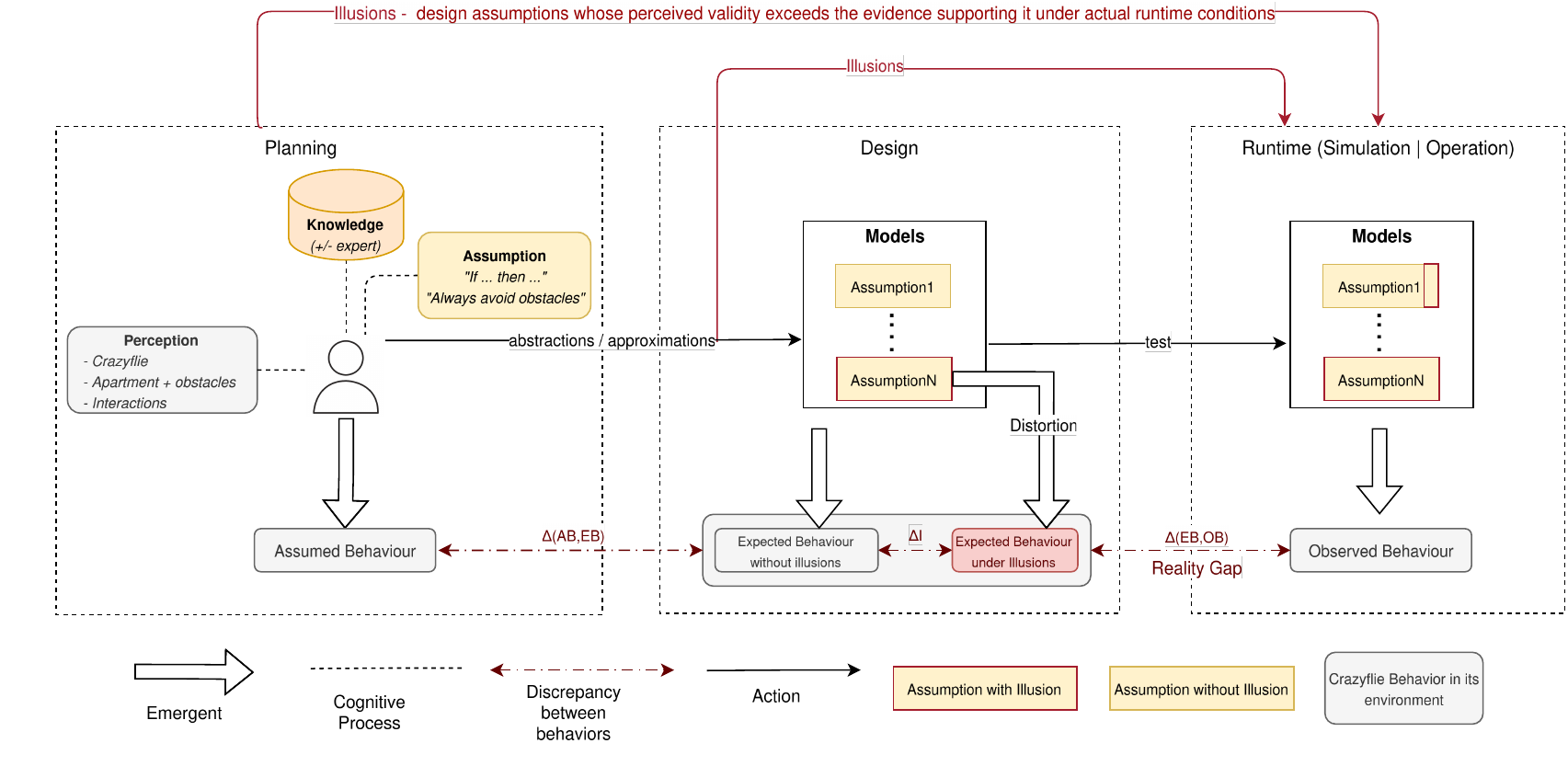}
    \caption{Problem Definition}
    \Description{Problem definition describing how illusions evolves during conception's life cycle}
    \label{fig:problem_def}
\end{figure*}


As explained in the introduction, we define a \textit{design illusion} as the persistent reliance on invalidated assumptions and the resulting false confidence in the design model's operational validity. An assumption considered valid during the design phase may become invalid during operation because of modelling approximations, environment unpredictable nature, etc. This distinguishes an illusion from an implementation \textit{bug}: the program's actual behaviour deviates from its formal semantic specification.

The conception of a Cyber-Physical System unfolds through several phases before operation. We distinguish three main phases in the process: \textit{Planning}, \textit{Design}, and \textit{Runtime}, which encompasses both simulation and operation. 
Across these phases, the system's behaviour is successively conceptualised, formalised, and finally executed, giving rise to four distinct behaviours, defined below and illustrated in Figure~\ref{fig:problem_def}.

\paragraph{\textbf{Assumed behaviour (AB)} is the conceptual representation, made during stakeholders' cognitive process, of how the system is believed to behave within its environment before any formalization or implementation. It emerges from prior knowledge, perception, and assumptions regarding the system and its operational context. At this stage, the behaviour reflects stakeholders' belief which reflects a more or less deep understanding of the system and its environment}

\paragraph{In the context of Model-Driven Engineering (MDE),\textbf{Expected Behaviour (EB)}, is the formal, model-based representation specifying how the system is anticipated to behave within its environment after stakeholders' assumptions have been incorporated into design models. These models embed both \textit{explicit assumptions}, directly formulated by stakeholders, and \textit{implicit assumptions}, which are not necessarily known. At this stage, all underlying assumptions are considered valid and consistent with the intended operational context. This behaviour remains implicit and is not yet observable.}

\paragraph{\textbf{Expected Behaviour under illusions (EB-I)} corresponds to an Expected Behaviour for which one or more underlying assumptions may not hold in a given context, \textit{e.g.,} due to changes in the system, environment, or operational conditions. 
Consequently, the anticipated behaviour may be partially incorrect, overgeneralized, or insufficiently validated.}

\paragraph{\textbf{Observed Behaviour (OB)} can be measured when the system is in operation or simulated. It reflects the real or simulated interactions, uncertainties, and constraints encountered during execution. It constitutes the ground truth against which expected behaviour is compared, revealing discrepancies caused by modeling limitations, imperfect assumptions, or illusions.}

\paragraph{All these behaviours are connected through a sequence of shifts, where each one introduces potential discrepancies due to the increasing distance between the stakeholder's perception of reality and the actual system-environment interactions.}

The shift from \textbf{AB} to \textbf{EB} corresponds to the transformation of stakeholders' conceptual representation into a formal model. During this process, assumptions about the system and its environment are defined as specifications and integrated in the models, typically using abstractions and approximations. Since formal models necessarily rely on abstractions and approximations, they may not fully represent all aspects of the original perception of reality. Consequently , some information may be simplified, omitted, or transformed, leading to a discrepancy $\Delta(AB,EB)$ between the behaviour assumed by stakeholders and the expected implemented one in the model.

The shift from \textbf{EB} to \textbf{EB-I} occurs when one or more assumptions in the model are later found to be invalid under actual operational conditions. In this case, the model itself has not changed but design illusions, $\Delta(I)$, result from distortions implied by abstractions and approximations. 
The validity of an assumption, however, cannot be verified without being tested. It can be evaluated through targeted experiments including tests on individual components or subsystems, simulation which has its own assumptions, or system operation.

The shift from \textbf{EB} or \textbf{EB-I} to \textbf{OB} occurs when the system is executed within simulation or operation environment. At his stage, the system is exposed to external conditions, interactions not fully captured during design, and design illusions. The comparison between expected and observed behaviours reveals the \textit{Reality Gap}, denoted as $\Delta(EB,OB)$. This gap represents the observable consequence of the accumulated discrepancies between design representations and reality.  It can have a neutral, positive, or negative effect on the system and its environment, and it can grow over the system’s life cycle as illusions accumulate or as new ones emerge from the evolving interaction between the system and its environment.

\subsection{Toward an illusions-awareness pipeline}
\begin{figure*}
    \centering
    \includegraphics[width=0.9\linewidth]{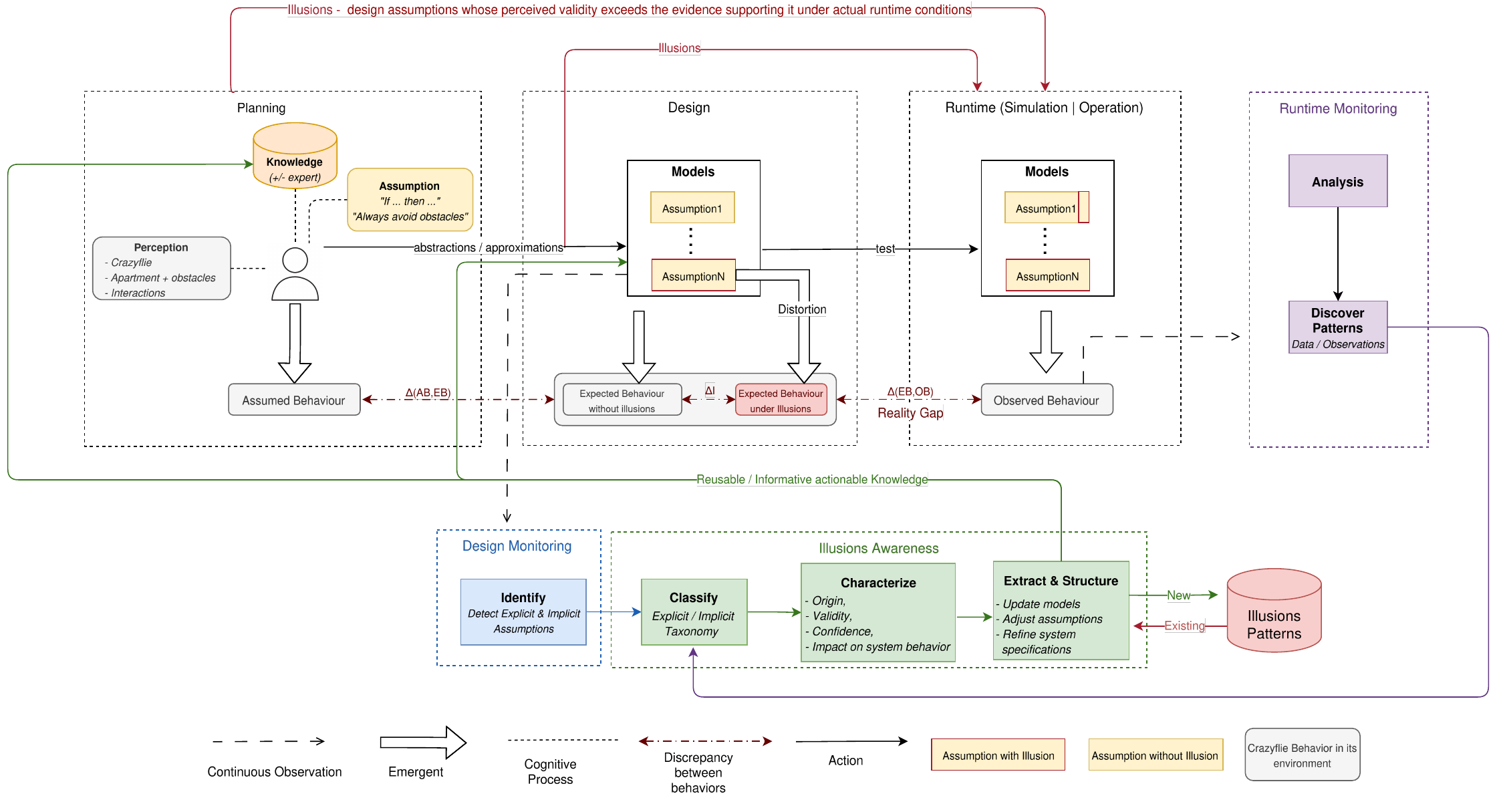}
    \caption{A Conceptual Pipeline for Design Illusions Awareness}
    \Description{Conceptual pipeline of illusions awareness}
    \label{fig:illusions_awareness}
\end{figure*}

The previous section discussed that design illusions cannot be entirely avoided during Planning and Design phases. Some assumptions will inevitably turn out to be invalid once the system is confronted with simulation or operation. Rather than attempting to eliminate such illusions entirely, the objective is to identify them, understand their causes, and exploit the knowledge they reveal. This knowledge can then be used to refine design models, increase stakeholders' awareness to support better design decisions. To this end, we propose a conceptual pipeline to identify, classify, characterize and leverage these illusions into reusable, informative knowledge for system designer, presented below and illustrated in Figure~\ref{fig:illusions_awareness}.

\subsubsection{Identify}

\textbf{Design monitoring}, aims at analyzing the models, code and parameters produced during Design. The objective is to automatically extract, for instance based on priori knowledge, the explicit and implicit assumptions on which the system relies--\textit{e.g.,} \textit{"The wall is always detectable"} or \textit{"distance measurements are accurate"}. 
Explicit assumptions are directly stated by developers; implicit ones are undocumented but embedded in the model design or implemented code--remaining challenging to extract automatically.

\textbf{Runtime monitoring} : analyzes data produced while the system is executing--sensor readings, internal states, control decisions--to detect patterns of failure or deviation from the expected behaviour. For instance, a sudden increase in the measured distance to the wall, or a repeated loss of wall tracking. In our example, such monitoring could reveal a recurring pattern : \textit{e.g.,} the drone consistently loses track of the wall when passing near an obstacle.
What can be observed is constrained by the probes deployed, typically (i) within the CPS to observe its internal state or its environment, or (ii) within the environment itself, observing both the environment and the CPS’s behaviour.


Design monitoring can identify candidate assumptions (as noted in Section 3.1), their validity can only be established through testing and experimentation. An illusion is thus identified when runtime monitoring reveals repeated deviations traceable to such a candidate, establishing both persistent reliance and the false confidence it produced.

\subsubsection{Classify}
Once identified, illusions can be classified making easier reasoning and decision-making. In particular, a taxonomy linking types of illusions and reality gaps to their possible sources would be highly beneficial (\textit{e.g.}, to reduce a reality gap, improve approximations). A first classification distinguishes explicit from implicit assumptions. A second one relates each assumption to the part of the system's architecture it concerns, \textit{e.g.,} perception (sensing), computation (decision-making), or actuation (control). 

\subsubsection{Characterize}
Classified assumptions are then characterized along several dimensions, in order to understand not only their nature, but also how and when they can be invalid.
The \textbf{Origin} of the assumption details how the assumption was obtained. For instance, it distinguishes (i) deductive reasoning such as rule-based reasoning, which is derived from an idealized model of the system or its environment from (ii) inductive reasoning inferred from previous flight experience or observed data. 
The \textbf{Validity} establishes the boundary conditions under which a given assumption remains true. It defines the specific operational contexts where the assumption holds valid and delineates the environmental or situational factors—such as high clutter, reflective surfaces, or low-light conditions—that may cause its invalidity.
The \textbf{Confidence} reflects how reliable an assumption is considered to be, based on the reasoning approach used to derive it, the coverage of the data supporting it, and the designer's own knowledge and experience. In practice, this confidence is also influenced by sensor quality and performance, as well as by the quality of the historical data used to build the assumption.
The \textbf{Impact} describes what happens to the system's behaviour if the assumption turns out to be false. For instance, the drone drifting away from the wall, colliding with an obstacle, or the mission failing altogether. Characterizing this impact also means identifying possible corrective responses, such as adjusting the trajectory, slowing down in an uncertain zone, switching to a 'safe exploration' mode, or stopping in a stationary hover.

\subsubsection{Leverage} 
Characterized illusions can then be structured into reusable knowledge that supports future design activities. This knowledge could be leveraged both autonomously by the system or manually by stakeholders. On the one hand, it can enrich the system's internal models by updating their representation. On the other hand, it provides informative feedback to designers, making explicit which assumptions repeatedly fail, under which contexts, and with what consequences.
Such feedback enables more informed future design decisions. It can be used to update design models, adjust or replace assumptions whose validity has been challenged, and refine system specifications by making previously implicit assumptions explicit or by introducing new operational constraints. Over time, accumulating knowledge from recurring design illusions could contributes to a continuously evolving knowledge base, \textit{i.e.,} illusions patterns in Figure~\ref{fig:illusions_awareness}, that improves both system robustness and the designers' understanding of illusions contributing to the reality gap.

\section{Related work}

Assumptions play a central role in the design and validation of complex Cyber-Physical Systems. In automated driving, identifying the assumptions underlying scenario-based testing is considered essential to ensure the validity of testing results \cite{neurohr2020}. Recent studies have also investigated assumption, by identifying multiple typical categories of assumptions in cyber-physical systems \cite{li2025}. However, how invalidated assumptions contribute to reality gaps and how this knowledge can be reused to support future design decisions remain insufficiently explored.

Complex autonomous systems rely on assumptions linking their intended purpose, operational context, and technical realization. Stellet et al. \cite{Stellet2019} introduced the notion of deductive gaps, using a three-circle model to show that violated assumptions can lead to discrepancies between required, specified, and implemented behaviours. While their approach focuses on detecting and validating assumptions to ensure system validity, it does not consider their invalidation as reusable design knowledge. In contrast, we identify a fourth category of discrepancy, corresponding to design illusions.

The reality gap has been extensively studied in robotics, and in particular through the concept of sim-to-real transfer. Aljalbout \textit{et al}. \cite{aljalbout2025realitygaproboticschallenges} survey methods to reduce or overcome it by improving simulation fidelity or making policies robust to unmodeled discrepancies. However, it treats the reality gap primarily as an engineering problem to be minimized, without characterizing the underlying invalid assumptions as reusable design knowledge.

The concept of reality gap has also been widely studied in the context of digital twins. Ma \textit{et al.} \cite{Ma2026} identify \textit{context mismatch} as arising when a digital twin's encoded assumptions no longer align with the physical asset's evolving operating context, which echoes our notion of design illusions. However, they address this mismatch through online recalibration, treating it as a correction problem rather than as reusable design knowledge. In contrast, our conceptual pipeline aims to exploit invalidated assumptions as a knowledge to support future design decisions.

\section{Conclusion}

Designing Cyber-Physical systems requires making assumptions about the system, its environment, and their interactions. While these assumptions are necessary to make design decisions tractable, they may become invalid when the system is confronted with real operating conditions. In this paper, we introduced the concept of \textit{design illusions} to describe situations where a system behave according to its design, but where one or more underlying design assumptions do not hold in the encountered context. Unlike implementation errors, design illusions reveal limitations in the designer's representation of reality.


Based on this concept, we proposed a conceptual pipeline to identify, classify, characterize, and leverage design illusions as reusable knowledge--combining runtime and design monitoring, assumption characterization, and feedback mechanisms to transform unexpected behaviours into insights supporting future design decisions.

Future work will focus on the formalization of design illusions and their integration into engineering processes. This includes developing automated approaches to extract and monitor design assumptions, defining metrics to evaluate their validity and impact, and investigating how the resulting knowledge can be autonomously leveraged by systems or used by designers to refine models, assumptions, and specifications. Ultimately, this approach could contribute to more adaptive and better-informed design processes for such complex systems.

\bibliographystyle{ACM-Reference-Format}
\bibliography{acmart}


\begin{thebibliography}{11}


\ifx \showCODEN    \undefined \def \showCODEN     #1{\unskip}     \fi
\ifx \showISBNx    \undefined \def \showISBNx     #1{\unskip}     \fi
\ifx \showISBNxiii \undefined \def \showISBNxiii  #1{\unskip}     \fi
\ifx \showISSN     \undefined \def \showISSN      #1{\unskip}     \fi
\ifx \showLCCN     \undefined \def \showLCCN      #1{\unskip}     \fi
\ifx \shownote     \undefined \def \shownote      #1{#1}          \fi
\ifx \showarticletitle \undefined \def \showarticletitle #1{#1}   \fi
\ifx \showURL      \undefined \def \showURL       {\relax}        \fi
\providecommand\bibfield[2]{#2}
\providecommand\bibinfo[2]{#2}
\providecommand\natexlab[1]{#1}
\providecommand\showeprint[2][]{arXiv:#2}

\bibitem[Acker et~al\mbox{.}(2024)]%
        {Acker2024}
\bibfield{author}{\bibinfo{person}{Bert~Van Acker}, \bibinfo{person}{Paul~De
  Meulenaere}, \bibinfo{person}{Hans Vangheluwe}, {and}
  \bibinfo{person}{Joachim Denil}.} \bibinfo{year}{2024}\natexlab{}.
\newblock \showarticletitle{Validity Frame–enabled model-based engineering
  processes}.
\newblock \bibinfo{journal}{\emph{SIMULATION}} \bibinfo{volume}{100},
  \bibinfo{number}{2} (\bibinfo{year}{2024}), \bibinfo{pages}{185--226}.
\newblock
\showeprint{https://doi.org/10.1177/00375497231205035}
\href{https://doi.org/10.1177/00375497231205035}{doi:\nolinkurl{10.1177/00375497231205035}}


\bibitem[Aljalbout et~al\mbox{.}(2025)]%
        {aljalbout2025realitygaproboticschallenges}
\bibfield{author}{\bibinfo{person}{Elie Aljalbout}, \bibinfo{person}{Jiaxu
  Xing}, \bibinfo{person}{Angel Romero}, \bibinfo{person}{Iretiayo Akinola},
  \bibinfo{person}{Caelan~Reed Garrett}, \bibinfo{person}{Eric Heiden},
  \bibinfo{person}{Abhishek Gupta}, \bibinfo{person}{Tucker Hermans},
  \bibinfo{person}{Yashraj Narang}, \bibinfo{person}{Dieter Fox},
  \bibinfo{person}{Davide Scaramuzza}, {and} \bibinfo{person}{Fabio Ramos}.}
  \bibinfo{year}{2025}\natexlab{}.
\newblock \bibinfo{title}{The Reality Gap in Robotics: Challenges, Solutions,
  and Best Practices}.
\newblock
\showeprint[arxiv]{2510.20808}~[cs.RO]
\urldef\tempurl%
\url{https://arxiv.org/abs/2510.20808}
\showURL{%
\tempurl}


\bibitem[Broman and Woodcock(2023)]%
        {Broman2023}
\bibfield{author}{\bibinfo{person}{David Broman} {and} \bibinfo{person}{Jim
  Woodcock}.} \bibinfo{year}{2023}\natexlab{}.
\newblock \showarticletitle{What are the fundamental software abstractions for
  designing reliable cyber-physical systems operating in uncertain
  environments?}
\newblock \bibinfo{journal}{\emph{Research Directions: Cyber-Physical Systems}}
   \bibinfo{volume}{1} (\bibinfo{year}{2023}), \bibinfo{pages}{e4}.
\newblock
\href{https://doi.org/10.1017/cbp.2023.4}{doi:\nolinkurl{10.1017/cbp.2023.4}}


\bibitem[Derler and Lee(2012)]%
        {derler2012}
\bibfield{author}{\bibinfo{person}{Patricia Derler} {and}
  \bibinfo{person}{Edward Lee}.} \bibinfo{year}{2012}\natexlab{}.
\newblock \showarticletitle{Modeling Cyber-Physical Systems}.
\newblock \bibinfo{journal}{\emph{Proc. IEEE}}  \bibinfo{volume}{100}
  (\bibinfo{date}{01} \bibinfo{year}{2012}), \bibinfo{pages}{13--28}.
\newblock
\href{https://doi.org/10.1109/JPROC.2011.2160929}{doi:\nolinkurl{10.1109/JPROC.2011.2160929}}


\bibitem[Lee(2006)]%
        {Lee2006}
\bibfield{author}{\bibinfo{person}{Edward~A. Lee}.}
  \bibinfo{year}{2006}\natexlab{}.
\newblock \showarticletitle{Cyber-Physical Systems - Are Computing Foundations
  Adequate?}. In \bibinfo{booktitle}{\emph{NSF Workshop on Cyber-Physical
  Systems: Research Motivation, Techniques and Roadmap}}.
  \bibinfo{address}{Austin, TX}.
\newblock
\newblock
\shownote{Position Paper}.


\bibitem[Li et~al\mbox{.}(2025)]%
        {li2025}
\bibfield{author}{\bibinfo{person}{Chengyu Li}, \bibinfo{person}{Saleh
  Faghfoorian}, {and} \bibinfo{person}{Ivan Ruchkin}.}
  \bibinfo{year}{2025}\natexlab{}.
\newblock \bibinfo{title}{What Does It Take to Get Guarantees? Systematizing
  Assumptions in Cyber-Physical Systems}.
\newblock
\showeprint[arxiv]{2511.15952}~[eess.SY]
\urldef\tempurl%
\url{https://arxiv.org/abs/2511.15952}
\showURL{%
\tempurl}


\bibitem[Ma et~al\mbox{.}(2026)]%
        {Ma2026}
\bibfield{author}{\bibinfo{person}{Sizhe Ma}, \bibinfo{person}{Katherine~A.
  Flanigan}, {and} \bibinfo{person}{Mario Bergés}.}
  \bibinfo{year}{2026}\natexlab{}.
\newblock \showarticletitle{Bridging the Reality Gap in Digital Twins with
  Context-Aware, Physics-Guided Deep Learning}.
\newblock \bibinfo{journal}{\emph{Journal of Computing in Civil Engineering}}
  \bibinfo{volume}{40}, \bibinfo{number}{3} (\bibinfo{date}{May}
  \bibinfo{year}{2026}).
\newblock
\showISSN{1943-5487}
\href{https://doi.org/10.1061/jccee5.cpeng-7024}{doi:\nolinkurl{10.1061/jccee5.cpeng-7024}}


\bibitem[Marti et~al\mbox{.}(2019)]%
        {Marti2019}
\bibfield{author}{\bibinfo{person}{Enrique Marti},
  \bibinfo{person}{Miguel~Angel de Miguel}, \bibinfo{person}{Fernando Garcia},
  {and} \bibinfo{person}{Joshué Pérez~Rastelli}.}
  \bibinfo{year}{2019}\natexlab{}.
\newblock \showarticletitle{A Review of Sensor Technologies for Perception in
  Automated Driving}.
\newblock \bibinfo{journal}{\emph{IEEE Intelligent Transportation Systems
  Magazine}}  \bibinfo{volume}{PP} (\bibinfo{date}{09} \bibinfo{year}{2019}),
  \bibinfo{pages}{1--1}.
\newblock
\href{https://doi.org/10.1109/MITS.2019.2907630}{doi:\nolinkurl{10.1109/MITS.2019.2907630}}


\bibitem[Neurohr et~al\mbox{.}(2020)]%
        {neurohr2020}
\bibfield{author}{\bibinfo{person}{Christian Neurohr}, \bibinfo{person}{Lukas
  Westhofen}, \bibinfo{person}{Tabea Henning}, \bibinfo{person}{Thies de
  Graaff}, \bibinfo{person}{Eike Möhlmann}, {and} \bibinfo{person}{Eckard
  Böde}.} \bibinfo{year}{2020}\natexlab{}.
\newblock \bibinfo{title}{Fundamental Considerations around Scenario-Based
  Testing for Automated Driving}.
\newblock
\showeprint[arxiv]{2005.04045}~[cs.SE]
\urldef\tempurl%
\url{https://arxiv.org/abs/2005.04045}
\showURL{%
\tempurl}


\bibitem[Rocher(2025)]%
        {crazyflieGithub}
\bibfield{author}{\bibinfo{person}{Gerald Rocher}.}
  \bibinfo{year}{2025}\natexlab{}.
\newblock \bibinfo{title}{Crazyflie2.1 Github Project}.
\newblock
\urldef\tempurl%
\url{https://github.com/gerald-rocher/Crazyflie2.1}
\showURL{%
\tempurl}


\bibitem[Stellet et~al\mbox{.}(2019)]%
        {Stellet2019}
\bibfield{author}{\bibinfo{person}{Jan~Erik Stellet}, \bibinfo{person}{Tino
  Brade}, \bibinfo{person}{Alexander Poddey}, \bibinfo{person}{Stefan
  Jesenski}, {and} \bibinfo{person}{Wolfgang Branz}.}
  \bibinfo{year}{2019}\natexlab{}.
\newblock \showarticletitle{Formalisation and algorithmic approach to the
  automated driving validation problem}. In \bibinfo{booktitle}{\emph{2019 IEEE
  Intelligent Vehicles Symposium (IV)}}. \bibinfo{pages}{45--51}.
\newblock
\href{https://doi.org/10.1109/IVS.2019.8813894}{doi:\nolinkurl{10.1109/IVS.2019.8813894}}


\end{thebibliography}
\end{document}